\documentclass{article}

\usepackage{arxiv}

\usepackage[utf8]{inputenc} 
\usepackage[T1]{fontenc}    
\usepackage{hyperref}       
\usepackage{url}            
\usepackage{booktabs}       
\usepackage{amsfonts}       
\usepackage{nicefrac}       
\usepackage{microtype}      
\usepackage{lipsum}		
\usepackage{graphicx}
\usepackage{natbib}
\usepackage{doi}
\usepackage{float}
\usepackage{multirow}%
\usepackage{amsmath,amssymb,amsfonts}%
\usepackage{amsthm}%
\usepackage{mathrsfs}%
\usepackage[title]{appendix}%
\usepackage{xcolor}%
\usepackage{textcomp}%
\usepackage{manyfoot}%
\usepackage{booktabs}%

\usepackage{algorithm}%
\usepackage{algorithmicx}%
\usepackage{algpseudocode}%
\usepackage{listings}%

\usepackage{makecell} 
\usepackage{mathtools}

\usepackage[acronym]{glossaries}
\newacronym{iou}{IoU}{Intersection over Union}
\newacronym{sde}{SDE}{stochastic differential equation}
\newacronym{vae}{VAE}{variational autoencoder}
\newacronym{ddpm}{DDPM}{denoising diffusion probabilistic model}
\newacronym{mcmc}{MCMC}{Markov chain Monte Carlo}
\newacronym{glas}{GlaS}{Gland Segmentation}
\newacronym{sdf}{SDF}{signed distance function}
\newacronym{gan}{GAN}{generative adversarial network}
\newacronym{map}{MAP}{maximum a posteriori}
\newacronym{mmse}{MMSE}{minimum mean square error}
\newacronym{cnn}{CNN}{convolutional neural network}
\newacronym{cnns}{CNNs}{convolutional neural networks}
\newacronym{he}{H\&E}{Haematoxylin and Eosin}
\newacronym{ema}{EMA}{exponential moving average}
\newacronym{ve}{VE}{variance-exploding}
\newacronym{vp}{VP}{variance-preserving}
\newacronym{ldm}{LDM}{latent diffusion model}
\newacronym{cfm}{CFM}{conditional flow matching}
\newacronym{nfe}{NFE}{number of function evaluations}
\newacronym{cnf}{CNF}{continuous normalizing flow}
\newacronym{nf}{NF}{normalizing flow}
\newacronym{fm}{FM}{flow matching}
\newacronym{ode}{ODE}{ordinary differential equation}

\hypersetup{
  colorlinks,
  citecolor=black,
  linkcolor=black,
  urlcolor=black}

\renewcommand{\d}[1]{\ensuremath{\operatorname{d}\!{#1}}}

\newcommand{\domainmasksdf}{\widetilde{\mathcal{M}}}
\newcommand{\domainimage}{\mathcal{X}}
\newcommand{\domainmask}{\mathcal{M}}
\newcommand{\minus}{\scalebox{0.75}[1.0]{$-$}}
\usepackage{siunitx}
\title{FlowSDF: Flow Matching for Medical Image Segmentation Using Distance Transforms}

\author{Lea Bogensperger \\
	Institute of Computer Graphics and Vision\\
	Graz University of Technology\\
	Graz, Austria\\
	\texttt{lea.bogensperger@tugraz.at}\\
	\And
	Dominik Narnhofer \\
	Photogrammetry and Remote Sensing\\
	ETH Zurich\\
	Zurich, Switzerland \\
	\texttt{dnarnhofer@ethz.ch} \\
	\AND
	Alexander Falk \\
	Institute of Computer Graphics and Vision\\
	Graz University of Technology\\
	\texttt{falk@tugraz.at} \\
	\And
	Konrad Schindler \\
	Photogrammetry and Remote Sensing\\
	ETH Zurich\\
	\texttt{schindler@ethz.ch} \\
	\And
	Thomas Pock \\
	Institute of Computer Graphics and Vision\\
	Graz University of Technology\\
	\texttt{thomas.pock@tugraz.at} \\
}

\renewcommand{\shorttitle}{\textit{arXiv} Template}

\hypersetup{
pdftitle={A template for the arxiv style},
pdfsubject={q-bio.NC, q-bio.QM},
pdfauthor={David S.~Hippocampus, Elias D.~Striatum},
pdfkeywords={flow matching, image segmentation, conditional generative models, signed distance function},
}

\begin{document}

\maketitle

\begin{abstract}
Medical image segmentation plays an important role in accurately identifying and isolating regions of interest within medical images. 
Generative approaches are particularly effective in modeling the statistical properties of segmentation masks that are closely related to the respective structures.
In this work we introduce FlowSDF, an image-guided conditional flow matching framework, designed to represent the \gls{sdf}, and, in turn, to represent an implicit distribution of segmentation masks.
The advantage of leveraging the \gls{sdf} is a more natural distortion when compared to that of binary masks.
Through the learning of a vector field associated with the probability path of conditional \gls{sdf} distributions, our framework enables accurate sampling of segmentation masks and the computation of relevant statistical measures. 
This probabilistic approach also facilitates the generation of uncertainty maps represented by the variance, thereby supporting enhanced robustness in prediction and further analysis.
We qualitatively and quantitatively illustrate competitive performance of the proposed method on a public nuclei and gland segmentation data set, highlighting its utility in medical image segmentation applications. 
\end{abstract}


\section{Introduction}\label{sec:intro}
\renewcommand{\thefootnote}{}

Medical image segmentation is the process of assigning each pixel in an image to a semantic class, which is crucial for extracting valuable insights for diagnosis and treatment planing. \footnotetext{Source code at \url{https://github.com/leabogensperger/FlowSDF}}
Traditionally, this task was handled by methods that have now been largely superseded by deep neural networks. These networks are particularly effective due to their ability to learn discriminatively from large data sets in an end-to-end manner~\citep{ronneberger2015u,badrinarayanan2017segnet,wang2020axial}. In addition to these discriminative models, generative models like \glspl{gan} and \glspl{ddpm} have emerged for image segmentation with the advantage of learning the underlying statistics of segmentation masks conditioned on the respective images of interest~\citep{iqbal2022generative,xun2022generative,amit2021segdiff}. A very recently introduced generative approach is flow matching~\citep{lipman2022flow,liu2022flow}, which uses an \gls{ode} to model the probability flow that transforms a data distribution into a simple, tractable prior distribution.\\
\begin{figure}[t]
    \centering 
    \includegraphics[width=\textwidth]{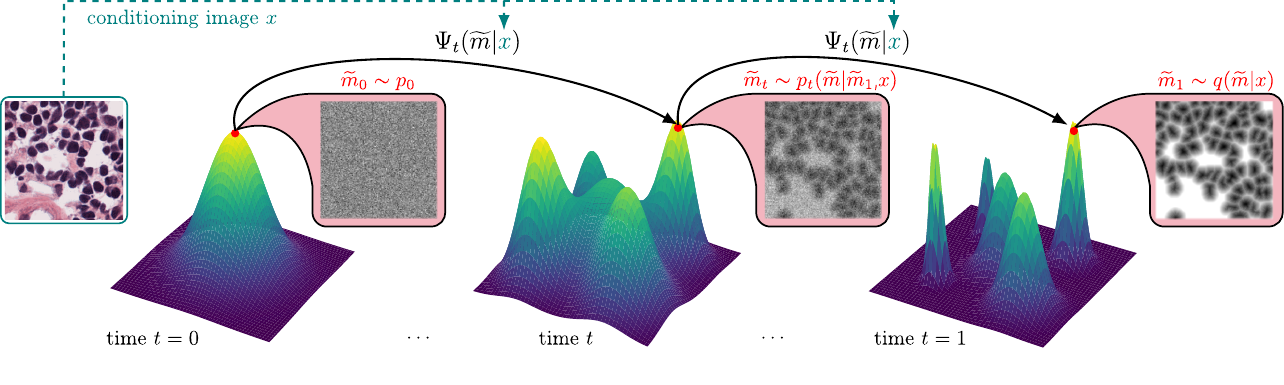}
    \caption{Overview of FlowSDF, which models a conditional flow $\Psi_t(\widetilde{m}|x)$ by relating a learned conditional vector field $v_{\theta}$ to the conditional probability path $p_t(\widetilde{m}|\widetilde{m}_1,x)$. During inference, a sample $\widetilde{m}_0\sim p_0$ is drawn from a known prior distribution (left) and the \gls{ode} is numerically solved to obtain segmentation masks $\widetilde{m}_1$ that resemble those from the conditional distribution $q(\widetilde{m}|x)$ (right).}
    \label{fig:overview}
\end{figure}
Furthermore, the process of modeling a probability flow can be extended to modeling the flow of a conditional probability, such as that of a segmentation mask given an image $p(m|x)$.
When modeling the conditional distribution using diffusion/flow models, noise is injected on the segmentation masks to transform them into samples from a Gaussian prior distribution. 
The statistical properties of these masks are complex -- typically bimodal or with only few modes based on the number of semantic classes involved.
It can be challenging to learn these statistics, as there is no transition between class modes, which can result in unnatural distortions.\\\\
A promising approach to mitigate these issues utilizes implicit segmentation maps, represented by \glspl{sdf}. An \gls{sdf} at any point on the segmentation map computes the orthogonal distance to the nearest boundary, assigning a negative sign to points inside the boundary and a positive sign to those outside. This method not only facilitates a smoother distribution of segmentation masks but also ensures more natural transitions between class modes.
Moreover, the use of \glspl{sdf} promotes smoothness in the segmentation maps, which are derived by thresholding the \gls{sdf} at object boundaries.\\\\
To this end, we introduce FlowSDF, an approach that combines flow matching with the \gls{sdf} for generative medical image segmentation. 
We build on prior work of~\citep{bogensperger2023score}
where a \gls{ve} \gls{sde} is employed to also model a conditional diffusion process to learn the statistics of segmentation masks given by the \gls{sdf}.
We improve upon our previous work in~\citep{bogensperger2023score} by using a generative flow matching framework~\citep{lipman2022flow} in place of the generative diffusion model. Our proposed image-guided flow matching method also learns a smooth implicit representation of segmentation masks conditioned on the corresponding images, see Figure~\ref{fig:overview}. 
This integration allows the noise-injection perturbation process to canonically smooth the distribution of \glspl{sdf} masks as further illustrated in Figure~\ref{fig:corruption_process}. 
Additionally, FlowSDF leverages its generative nature to produce multiple segmentations for a given input image, thereby allowing for segmentation uncertainty quantification. This also enhances the robustness and interpretability of our approach and provides valuable insights into the segmentation process. 
Summarized, our main contributions are:
\begin{itemize}
  \setlength\itemsep{-.25em}
    \item To the best of our knowledge, we are the first to apply conditional flow matching for image segmentation.
    \item Further, distance transforms are used to obtain a smoother, implicit representation of segmentation masks.
    \item Throughout multiple experiments we achieve competitive results for the considered data sets. 
\end{itemize}

\section{Related Work}

\subsection{Medical Image Segmentation}

In the following section, we discuss selected literature on medical image segmentation related to the proposed approach.
Historically, techniques such as level set methods have been prominent~\citep{osher1988fronts,osher2004level}, which use a signed function to compute distance metrics to the nearest object, whose actual contour is represented by the zero level. Hence, the aim is to provide a precise spatial context of the object of interest by focusing on the respective boundary. 
Another notable method in the field of variational methods is the Potts model, also known as piecewise constant Mumford-Shah model~\citep{chan2001active,mumford1989optimal}. It has been used extensively to handle segmentations by minimizing an energy function that discourages discontinuities between neighboring pixels unless there is a strong gradient in the image data, effectively enforcing smoothness while adapting to inherent image contours.
Undoubtedly, there are many more classical techniques for image segmentation, such as active contours~\citep{kass1988snakes} and graph cuts~\citep{boykov2004experimental,boykov2006graph}, to name but a few.\\\\
The emergence of deep learning led to many discriminative methods that have significantly advanced the field of image segmentation using deep neural networks~\citep{ronneberger2015u,badrinarayanan2017segnet}. These methods are designed to work end-to-end and have benefited greatly from continual advancements in architecture search, allowing for more refined and effective models~\citep{wang2020axial,isensee2021nnu,cao2022swin}.  
However, also some of these newer methods have recognized the vital information that can be extracted from information such as contours and boundaries, in particular for image segmentation. Therefore, there have been approaches incorporating shape priors and distance transforms for discriminative image segmentation using \glspl{cnn}~\citep{naylor2018segmentation,xue2020shape,brissman2021predicting,diff_augmentation_nuclei}. 
Although these discriminative methods tend to be fast, easy to deploy, and yield good results, they often exhibit overconfidence in their predictions, leading to issues with calibration where the predicted probabilities do not align with actual outcomes. 
\\\\
Furthermore, when annotation data is available from multiple annotators to account for ambiguity in segmentation data, discriminative models can be transformed into probabilistic variants by incorporating stochastic information by sampling from a learned latent distribution using a conditional \gls{vae}~\cite{kohl2018probabilistic,bhat2022generalized}. Several works build on this approach by utilizing the inherent hierarchical structure of U-Nets to integrate stochasticity~\cite{baumgartner2019phiseg,kohl2019hierarchical}. 
Ambiguous data labels can also be exploited by generative models, such as diffusion models~\cite{rahman2023ambiguous}; however, this ambiguity is often not explicitly present in many data sets.
\\\\
In contrast, generative models for image segmentation such as \glspl{gan}~\citep{isola2017image,li2021semantic} and score-based generative models~\citep{baranchuk2021label,amit2021segdiff,wolleb2022diffusion,wu2022medsegdiff,wu2024medsegdiff} have been utilized to focus on learning the data distributions of segmentation masks given input images. While \glspl{gan} are based on the principle of learning a direct mapping from noise to data through adversarial training, score-based diffusion models learn the log-gradient (score) of a data distribution by leveraging the idea of \glspl{sde} to gradually reverse a noise process. In our previous work, we use a score-based generative model where segmentation masks of corresponding medical images are represented implicitly by using \glspl{sdf}~\citep{bogensperger2023score}. By capturing the underlying data distributions, these models provide more robust predictions, especially in scenarios with complex or unseen data. Although they offer statistical robustness, they generally require more complex training processes and tend to operate slower.\\\\
A more recent generative method is flow matching~\citep{lipman2022flow,liu2022flow}, which extends \glspl{cnf} by a simulation-free training strategy. This method learns a vector field related to a conditional probability path, establishing a connection with diffusion-based methods. 
Likewise, the goal is to transition from a distribution of interest to a simple, tractable prior distribution – e.g. by gradually injecting noise to smooth a data distribution. Flow matching has been shown to be a more robust and stable alternative to diffusion models offering enhanced performance~\citep{lipman2022flow}, and has recently been applied to high-quality, unconditional image synthesis~\citep{esser2024scaling}.
Moreover, recent studies demonstrated its effectiveness in various domains such as depth estimation, label-conditioned image generation and image inpainting~\citep{dao2023flow,gui2024depthfm,yariv2023mosaic}. The subsequent section will introduce flow matching in more detail.

\subsection{Flow Matching}\label{sec:related:fm}
\Glspl{cnf}~\citep{chen2018neural} characterize the target density \(q = [\Psi]_\sharp p_0\) as push-forward of a tractable initial density \(p_0\) under a deterministic transformation \(\Psi \colon \mathbb R^d \to \mathbb R^d\). Typically, the transformation \(\Psi \colon \mathbb R^d \to \mathbb R^d\) is learned via maximum likelihood and maps samples \(x \sim p_0\) from the initial distribution to samples \(\Psi(x) \sim q\) that follow the target~\citep{rezende2015variational}. However, \Glspl{cnf} do not directly model \(\Psi\), but rather consider the temporal dynamics of a time-dependent transformation \(\Psi_t \colon [0, 1] \times \mathbb R^d \to \mathbb R^d\) with \(t \in [0, 1]\). The corresponding neural \gls{ode}~\citep{chen2018neural} is given as
\begin{equation}\label{eq:ode}
    \frac{\d{}}{\d t} \Psi_t(x) = v_{\theta,t}(\Psi_t(x)),
\end{equation}
where \(v_{\theta,t} \colon [0,1] \times \mathbb R^d \to \mathbb R^d\) is a parameterized network. It defines a time-dependent vector field by assigning spatial displacements to each spatio-temporal pair \((t, \Psi_t(x))\). The overall transformation \(\Psi(x) = \Psi_1(x) = x + \int_{0}^{1} v_{\theta,t}(x) \d{t}\) for the initial state \(x\) is then given by integrating the spatial displacements over the whole time domain, which can be done by utilizing off-the-shelf \gls{ode} solvers.\\
However, the widespread adoption of \glspl{cnf} as generative models has so far been hindered by the cost of maximum likelihood training. Because of this limitation, flow matching has been introduced as a feasible alternative~\citep{lipman2022flow,liu2022flow}. The essential insight is that any time-dependent vector field \(u_t \colon [0,1] \times \mathbb R^d \to \mathbb R^d\) which is related to the temporal dynamics of a given density path \(p_t \colon [0,1] \times \mathbb R^d \to [0, \infty)\) via the ubiquitous continuity equation~\citep{villani2009optimal} 
\begin{equation}\label{eq:continuity-equation}
   \frac{\d{}}{\d t} p_t(x) + \operatorname{div}( p_t(x)u_t(x) ) = 0,
\end{equation}
can be used as a regression target. The corresponding flow matching objective is thus given as
\begin{equation}\label{eq:fm-loss}
    \mathcal{L}_{\mathrm{FM}}(\theta) := \mathbb E_{t \sim \mathcal U_{[0, 1]}}\mathbb E_{x \sim p_t}\bigl[\tfrac 1 2 \Vert v_{\theta,t}(x) - u_t(x)\Vert^2\bigr].
\end{equation}
This objective is justified by identifying the density path in Eq.~\eqref{eq:continuity-equation} with the one that is induced by the continuous transformation, i.e., \(p_t = [\Psi_t]_\sharp p_0\). Thus, it becomes evident that minimizing Eq.~\eqref{eq:fm-loss} indeed yields a vector field \(v_{\theta,t}(\cdot)\) characterizing \(\Psi_t\) via Eq.~\eqref{eq:ode}. However, as the marginal ground truth vector field \(u_t\) is not known in general and \(p_t\) can take a complicated form, directly minimizing Eq.~\eqref{eq:fm-loss} is impossible. Instead, Lipman~et~al.~\citep{lipman2022flow} show that \(u_t\) can be constructed by superimposing many conditional vector fields \(u_t(\cdot | x_1)\) that depend on available training samples \(x_1 \sim q\). The simpler conditional flow matching objective is thus given as
\begin{equation}\label{eq:cfm-loss}
    \mathcal{L}_{\mathrm{CFM}}(\theta) = \mathbb E_{t \sim \mathcal U_{[0, 1]}} \mathbb E_{x_1 \sim q} \mathbb E_{x \sim p_t(\cdot | x_1)}\bigl[\tfrac 1 2 \Vert v_{\theta,t}(x) - u_t(x| x_1)\Vert^2\bigr],
\end{equation}
which yields the same set of minimizers as Eq.~\eqref{eq:fm-loss} and reveals similarities to score-based optimization~\citep{hyvarinen2005estimation}. To obtain an explicit expression for the regression target \(u_t(x | x_1)\), they propose conditional affine Gaussian flows \(\Psi_t(x | x_1) = \sigma_t x + \mu_t(x | x_1)\). According to Eq.~\eqref{eq:ode} it is of the form \(u_t(x | x_1)=\frac{x - \mu_t(x | x_1)}{\sigma_t}\frac{\d{}}{\d t}\sigma_t + \frac{\d{}}{\d t}\mu_t(x | x_1)\). In addition, the corresponding density path has the form \(p_t(x | x_1) = \mathcal{N}\bigl(x | \mu_t(x | x_1), \sigma_t^2 \operatorname{Id}\bigr)\). 
Following~\citep{lipman2022flow}, it starts from the tractable multivariate standard normal distribution \(p_0(\cdot | x_1) = \mathcal{N}\bigl(\cdot | 0, \operatorname{Id}\bigr)\) and ends in a Dirac peak at the conditioning sample \(x_1\), i.e., \(p_1(\cdot | x_1) = q(\cdot | x_1) = \mathcal{N}\bigl(\cdot | x_1, \sigma_{\mathrm{min}}^2 \operatorname{Id}\bigr)\), \(\sigma_{\mathrm{min}}^2 \approx 0\). In contrast, the temporal interpolation between \((\mu_0, \sigma_0)\) and \((\mu_1, \sigma_1)\)  is not uniquely defined and offers some degree of freedom. In this work, we adopt the best-performing variant from~\cite{lipman2022flow} that induces conditional optimal transport paths via \(\mu_t(x|x_1) = t x_1\) and \(\sigma_t = 1 - (1 - \sigma_{\mathrm{min}})t\).

\section{Method}

\subsection{Image Segmentation using \glspl{sdf}}\label{sec:method:seg_sdf}

\begin{figure}[t]
    \centering
    \includegraphics[width=\textwidth]{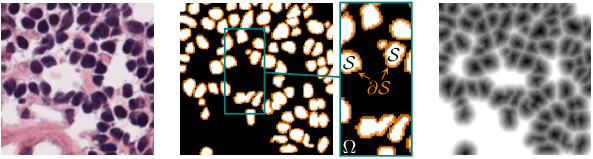}
    \caption{Given an image $x$ (left), its binary segmentation mask (center) can be transformed into a truncated \gls{sdf} mask (right). The zoomed area shows some segmentation objects in detail denoted by $\mathcal{S}$ and their boundaries $\partial \mathcal{S}$ embedded in the domain $\Omega$.}
    \label{fig:sdf_segm}
\end{figure}

Let $M,N$ be the spatial dimensions of the discretized images with domain $\Omega=\{1,\dots,M\}\times\{1,\dots,N\}$. We usually denote an image $x\in \domainimage := \mathbb{R}^{M\times N}$. For $C$-class semantic image segmentation the images have corresponding segmentation masks $m \in \domainmask$, where $\domainmask \coloneqq \{0,\dots,C-1\}^{M \times N}$, with $C=2$ for binary segmentation. Moreover, we consider the \gls{sdf} representation of a segmentation mask $\widetilde{m} \in \domainmasksdf$, which yields a continuous representation such that $\domainmasksdf \coloneqq \mathbb{R}^{M \times N}$.\\\\
Using \glspl{sdf} the segmentation problem can be rephrased in the context of Using \glspl{sdf}, the segmentation problem can be reformulated in terms of signed distances relative to the object boundaries. For each pixel $m_{ij}$ in the domain $\Omega$ with the object $\mathcal{S}$ to be segmented, the \gls{sdf} map assigns the distance to the nearest boundary pixel $\partial \mathcal{S}$. Additionally, the distance is signed, such that a negative sign indicates that the pixel is inside the object, while a positive sign indicates that it lies outside.
\\\\
The \gls{sdf} $\widetilde{m} \in \domainmasksdf$ of a segmentation mask $m$ can be computed by applying the Euclidian distance function to each pixel $m_{ij}$. To focus on the region near the object boundary $\partial \mathcal{S}$, the distances can be truncated at a threshold $\delta$. This truncation also prevents distant background pixels from influencing the segmentation outcome, as large positive distances are ignored. The implicit, truncated \gls{sdf} $\phi(m_{ij})$ is then given by (also see~\citep{osher2004level}):
\begin{align}
\phi(m_{ij}) = \begin{cases}
 \minus \min \{ \min_{y\in \partial S} \Vert y-m_{ij}\Vert_2, \delta \} \quad &\text{ if } m_{ij} \in \mathcal{S},\\
\min \{ \min_{y\in \partial S} \Vert y-m_{ij}\Vert_2, \delta \} \quad &\text{ if } m_{ij} \in \Omega \setminus \mathcal{S}, \\
0 \quad &\text{ if } m_{ij} \in \partial \mathcal{S}.
\end{cases}
\end{align}\\
Thus we can obtain the full segmentation map $\widetilde{m}=\phi(m_{ij})_{\substack{i=1,\dots,M \\ j=1,\dots,N}}$ by applying $\phi(\cdot)$ on the entire segmentation mask $m$, which is also depicted in Figure~\ref{fig:sdf_segm}. 
Conversely, given $\widetilde{m}$ one can easily retrieve the binary segmentation map $m$ by thresholding at 0 to separate segmented objects from background, as given by
\begin{equation}
    m = \mathbf{1}_{\{ \widetilde{m} \leq 0\}},
\end{equation}
where $\mathbf{1}_{\{ \cdot \}}$ denotes the indicator function that evaluates to 1 or 0 depending on whether the respective condition is fulfilled. 

\subsection{Motivating the \gls{sdf}}

\begin{figure}[t]
    \centering
    \includegraphics[width=\textwidth]{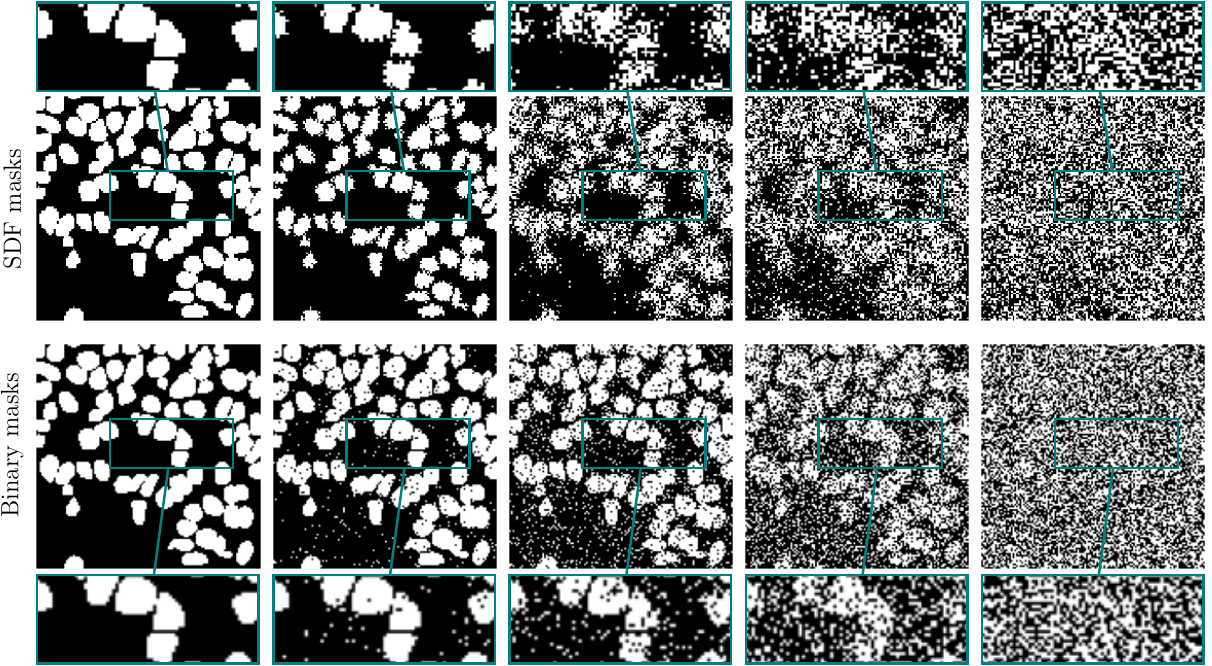}
    \caption{Comparison of the effect of the transitioning process on the resulting thresholded segmentation masks when using an \gls{sdf} mask $\widetilde{m}$ and a binary mask $m$ for a given image $x$. Note that the \gls{sdf} representation allows for a more natural distortion process in its thresholded masks, which evolves along the object boundaries instead of directly introducing ``hole''-like structures at random pixel positions as it is the case in the thresholded masks when directly using the binary segmentation mask.}
    \label{fig:sdf_comparison}
\end{figure}

In a naive approach, the push forward function $p_t = [\Psi_t]_\sharp p_0$ transitions the distribution of binary masks towards a tractable multivariate Gaussian by incrementally adding noise to the desired segmentation mask. However, the intermediate masks $m_t\sim p_t(m|m_1)$ resulting from this process suffer from arbitrary pixel-wise perforations.
This degradation occurs due to the discontinuous nature of segmentation masks. 
If additive noise causes the decision boundary to be surpassed, i.e. leading to a class jump, it results in arbitrary holes in the mask rather than the gradual removal of semantic information.\\
By transforming discontinuous masks using distance transforms, a smoother representation is provided while keeping the semantic information. 
In this setting the perturbation affects the masks starting from boundaries, subsequently progressing towards the center of objects until converging towards a sample from the tractable prior distribution. 
Therefore using \gls{sdf} representations of semantic masks introduces a more canonical way of converting the distribution of interest to the tractable prior. 
A comparison of the resulting transitioning process on both variants of segmentation mask representations for several intermediate time steps is shown in Figure~\ref{fig:sdf_comparison}. 

\subsection{Image-Guided Flow Matching} \label{sec:method:fm}
\begin{figure}[t]
    \centering
    \includegraphics[width=\textwidth]{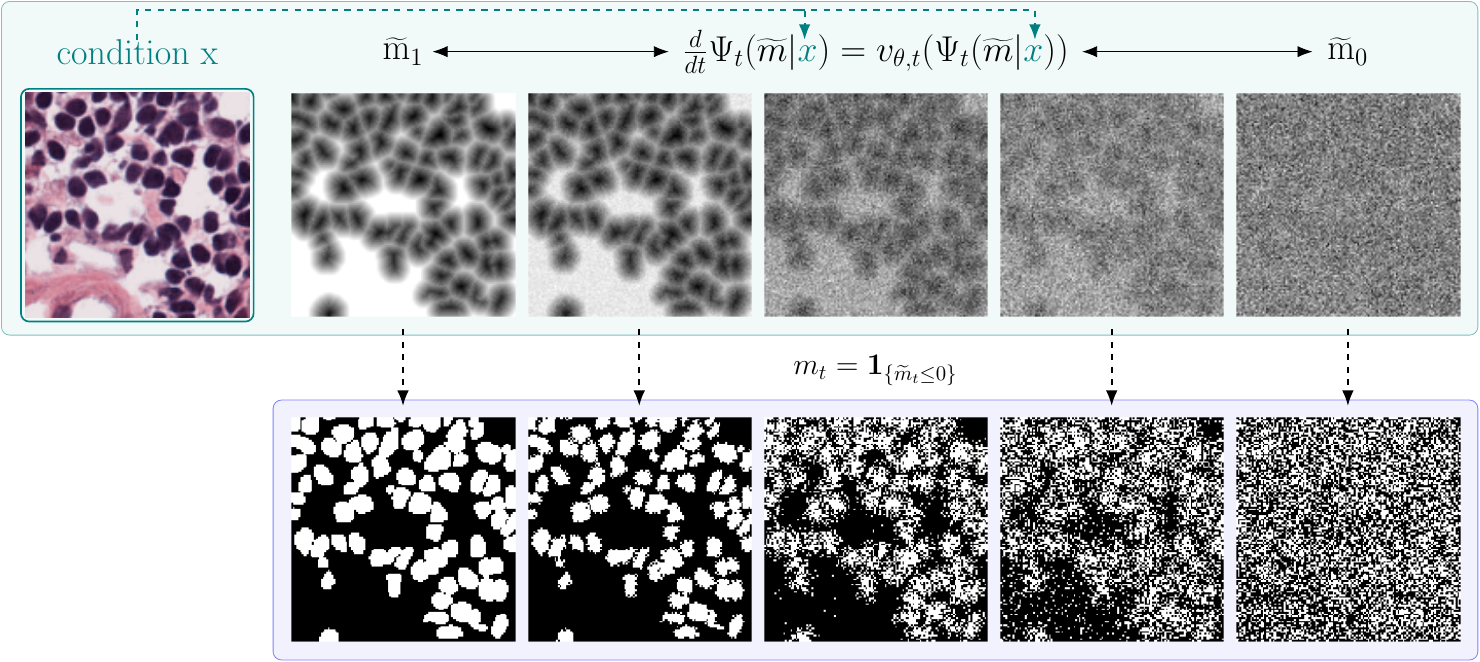}
    \caption{Schematic of the corruption/generative process (top row for different $t \in [0,1]$) on an \gls{sdf} segmentation mask $\widetilde{m}$ for a given image $x$. Note that the governing mechanism is bidirectional, hence to obtain samples during inference a numerical \gls{ode} integrator is used. 
   The corresponding binary segmentation masks $m_t$ are shown in the second row. While they are not directly used neither during training nor sampling, they can be obtained easily by thresholding and they demonstrate the induced corruption mechanism through the usage of the \gls{sdf}.}
    \label{fig:corruption_process}
\end{figure}

Given a data set of $S$ labeled pairs of images and segmentation masks $\mathcal{D}=\{(\widetilde{m}_s,x_s)\}_{s=1}^S$, the goal is to learn the conditional distribution of \gls{sdf}-represented segmentation masks given their respective images $q(\widetilde{m}|x)$.
Building on the introduced concept of flow matching from Section~\ref{sec:related:fm}, the aim is to learn a conditional vector field $v_{\theta,t}(\widetilde{m} |x)$ which is related to the sought conditional flow by (cf. Eq.~\eqref{eq:ode})
\begin{equation}\label{eq:ode-cond}
    \frac{\d{}}{\d t} \Psi_t(\widetilde{m}|x) = v_{\theta,t}(\Psi_t(\widetilde{m}|x),x).
\end{equation}
This is also depicted graphically in Figure~\ref{fig:corruption_process}, showing how the conditioning image $x$ guides the flow from the clean \gls{sdf} mask $\widetilde{m}_1$ to a corrupted version $\widetilde{m}_0$, which can be reversed by numerically integrating the \gls{ode}. Moreover, the corresponding thresholded binary segmentation masks $m_t$ during the different stages can be observed in the bottom row, however, note that they are only indirectly obtained while the image-guided flow matching explicitly works on the \gls{sdf}-transformed masks $\widetilde{m}_t$.
Then the final objective that is minimized is given by
\begin{equation}\label{eq:cfm-loss-cond}
    \mathcal{L}_{\mathrm{CFM}}(\theta) = \mathbb E_{t \sim \mathcal U_{[0, 1]}} \mathbb E_{(\widetilde{m}_1,x) \sim q(\widetilde{m}|x)} \mathbb E_{\widetilde{m}_t \sim p_t(\cdot | \widetilde{m}_1,x)}\bigl[\tfrac 1 2 \Vert v_{\theta,t}(\widetilde{m}_t,x) - u_t(\widetilde{m} | \widetilde{m}_1,x)\Vert^2\bigr].
\end{equation}
Learning the parameters $\theta$ of a neural network $v_{\theta,\cdot}$ such that it minimizes Eq.~\eqref{eq:cfm-loss-cond} is described in Algorithm~\ref{alg:Training}. Note that $v_{\theta,\cdot}(\cdot,x)$ additionally requires the conditioning image $x$ as an input. For the prior distribution $p(\widetilde{m}_0)$ a standard normal distribution $\mathcal{N}(0,\mathrm{Id})$ is used, although flow matching would also allow other choices.\\\\ 
\begin{table}
\begin{minipage}[t]{0.99\textwidth}
\begin{algorithm}[H]
    \caption{Image-Guided Flow Matching Training}
    \begin{algorithmic}
        \State \textbf{Input} data set $\mathcal{D} = \{(\widetilde{m}_s,x_s)\}^S_{s=1} \thicksim q(\widetilde{m}|x)$
        \Repeat
            \State Sample $(\widetilde{m}_1,x) \thicksim \mathcal{D}$
            \State Sample $t \thicksim \mathcal{U}_{[0,1]}$
            \State Sample $\widetilde{m}_0 \thicksim p(\widetilde{m}_0)$
            \State $\widetilde{m}_t \gets t \widetilde{m}_1 + (1-t) \widetilde{m}_0$
            \State $\dot{\widetilde{m}}_t \gets \widetilde{m}_1 - \widetilde{m}_0$
            \State $\theta^{k+1} = \texttt{Adam}(\theta^k,\nabla_{\theta^k} \frac 12 \Vert v_{\theta^k_t} (\widetilde{m}_t,x) - \dot{\widetilde{m}}_t \Vert_2^2)$ 
        \Until{convergence}
    \end{algorithmic}
\label{alg:Training}
\end{algorithm}
\end{minipage}
\end{table}
Generating new samples using flow matching involves integrating the \gls{ode} in Eq.~\eqref{eq:ode-cond}, the procedure is given in detail in Algorithm~\ref{alg:Inference}. In general, one has to choose the number of \gls{nfe} which corresponds to the discretization of the \gls{ode} with $\eta$. Moreover, to compare with sampling based on \glspl{sde}, $T$ noise injection steps can be incorporated to reintroduce noise into the denoised mask $\widetilde{m}_{\cdot}^1$ with a linear schedule. If $T=1$, the original \gls{ode}-based sampling as given in~\citep{lipman2022flow} is recovered. \\ \\
By employing a generative model, the goal shifts from learning a discriminative model that predicts segmentation masks in an end-to-end manner to a model that ideally captures the underlying statistics of the conditional distribution $q(\widetilde{m}|x)$. 
As the noise injected inference procedure resembles a stochastic sampling process, individual runs can be seen as samples $\widetilde{m}_k\sim p(\widetilde{m}|x)$.
Therefore, the \gls{mmse} can be approximated by $\mathbb{E}[\widetilde{m}| x] \approx \frac{1}{K} \sum_{k=1}^K \widetilde{m}_k$ with $K$ independent inference runs.
Furthermore, the variance $\mathbb{V}[\widetilde{m}| x] \approx \frac{1}{K} \sum_{k=1}^K ( \widetilde{m}_k-\mathbb{E}[\widetilde{m}| x])^2$ can be approximated and yields insightful information about classification uncertainty.

\begin{table}[t]
\begin{minipage}[t]{0.99\textwidth}
\begin{algorithm}[H]
    \caption{Image-Guided Sampling}
    \begin{algorithmic}
        \State trained  $v_{\theta,\cdot}$, conditioning image $x$, \gls{ode} discretization $\eta = \tfrac{1}{\mathrm{NFE}}$, noise injection steps $T \in \mathbb{N}$ with $\tau = \frac{1}{T}$
       \State $\widetilde{m}_0 \thicksim p({\widetilde{m}_0})$
        \For{$t =0,\dots,1-\tau$}
         
        \For{$l=t,\dots,1-\eta$}
                \State $\widetilde{m}_{t}^{l+\eta} \gets \texttt{ODEstep}(v_{\theta,l}(\widetilde{m}_{t}^{l},x))$
        \EndFor
        \State $z\sim \mathcal{N}(0,I)$
        \State $\widetilde{m}_{t+\tau}  \gets (t+\tau) \widetilde{m}^1_{t} + (1-t-\tau)z$
        \EndFor
        \State \Return $\widetilde{m}_1$
    \end{algorithmic}
\label{alg:Inference}
\end{algorithm}
\end{minipage}
\end{table}

\section{Experiments}
\subsection{Data Sets and Preprocessing}
For the experimental setting, we utilize two publicly available data sets. The first data set is MoNuSeg~\citep{kumar2017dataset}, which consists of 30 training images and 14 test images. Each image is of size $1000\times 1000$ and overall they contain more than 21,000 annotated nuclei in \gls{he} stained microscopic images. For data preprocessing we resort to a structure-preserving color normalization~\citep{vahadane2015structure} as the different organ sites yield considerable intensity variations in the data. Subsequently, all images were resized to $500\times 500$ following~\citep{valanarasu2021medical}. During training, random crops of $128\times 128$ were used with additional random horizontal and vertical flips for data augmentation.\\\\
As a second medical data set, the \gls{glas} data set~\citep{sirinukunwattana2017gland} was used. It consists of 85 training and 80 test \gls{he} stained microscopic images with annotated glands from colorectal cancer tissue. Again, a structure-preserving color normalization~\citep{vahadane2015structure} is used also here for data preprocessing, which is followed by resizing all training and test images to $128\times 128$ inspired by~\citep{valanarasu2021medical}. Data augmentation in training is done as with the MoNuSeg data set.

\subsection{Training Details}
The deep learning architecture to approximate the vector field $v_{\theta,\cdot}(\cdot,x)$ is based on work presented in~\citep{nichol2021improved,amit2021segdiff}. 
The mechanism to condition the learned vector field on the image $x$ are adopted from~\citep{amit2021segdiff}, where the extracted features of the noisy mask $\widetilde{m}_t$ are additively combined with the feature maps of the conditioning image $x$. Note that this differs from other commonly used conditioning mechanisms that concatenate the extracted features~\citep{ozdenizci2023restoring,ho2022cascaded,saharia2022palette}, but the extensive ablation study over the conditioning mechanism in~\citep{amit2021segdiff} shows superior performance.\\\\
During training, we utilized the Adam optimizer with a learning rate of 1e-4. The batch size was set to 16, and the models were trained for approximately 30k epochs for the first data set and nearly 40k epochs for the second. The average training time per epoch was 5.4 seconds for the MoNuSeg data set and 15.5 seconds for the \gls{glas} data set, using an NVIDIA Titan RTX GPU. Additionally, \gls{ema} with a decay factor of 0.9999 was applied throughout the training process.
The detailed training procedure of image-guided flow matching is described in Algorithm~\ref{alg:Training}.

\subsection{Evaluation}\label{sec:exp:eval}

After training we obtain a network  $v_{\theta,\cdot}$ that regresses a vector field, capable of gradually transitioning $p(\widetilde{m}_0)$ to $p(\widetilde{m}_1|x)$.
Consequently, Algorithm~\ref{alg:Inference} can be employed to synthesize new samples $\widetilde{m}_1$ given the conditioning images $x$ that have to be segmented.
The \texttt{ODEstep} is implemented in practice using numerical integration provided by~\citep{torchdiffeq}, requiring about 0.85 seconds to generate 4 samples with 4 \glspl{nfe}.
The number of \gls{nfe} is set to 4, as flow matching in general already yields remarkable results with only few \glspl{nfe}, which is advantageous compared to standard \glspl{ddpm}.
Moreover, averaging over a number of $K$ predictions to obtain an estimate for the \gls{mmse} was considered, where the respective ablation that investigates the choice of $K$ can be found in Figure~\ref{fig:ablation_K} for both data sets. For multiple noise injection runs $T$, we set $T=100$ to enable a fair comparison with the \gls{sde} sampler in~\citep{amit2021segdiff}.\\\\
As the image size was set to $128 \times 128$ to reduce computational demands, the inference procedure for the MoNuSeg data set consists of overlapping patches of each test image. Note that for \gls{glas} this was not necessary, as the entire images were resized to $128 \times 128$. Hence, an averaging strategy, similar to that proposed by~\citep{ozdenizci2023restoring}, was used to merge the overlapping sections within each image. Thereby, the average of the predictions over 2-4 different patch segments was taken to diminish the effect of cropped nuclei at the borders. This also quantitatively improved the evaluation results further, see Table~\ref{tab:ablation_patch_merge}.\\\\
The obtained segmentation masks are then evaluated using the standard metrics F1 score and \gls{iou}. 
We compare our method to commonly referred benchmark models, ensuring that both U-Net variants and attention/transformer mechanisms are considered. Moreover, we also consider~\citep{amit2021segdiff} using the publicly available source code to obtain a comparison with a generative, conditional \gls{ddpm} that predict standard binary segmentation masks in the sampling process. Finally, the approach in~\citep{bogensperger2023score} is also compared with, which is conditional diffusion model based on a variance-exploding \gls{sde} where also \gls{sdf} masks are predicted in the sampling process.

\subsection{Results}
Table~\ref{tab:results} shows quantitative results for both data sets with our method and comparison methods. To enable a fair comparison the benchmark results were taken from~\citep{valanarasu2021medical} where possible, apart from SegDiff~\citep{amit2021segdiff}. If not specified, the number of additional noise injection steps $T=1$ is used with FlowSDF.
The results indicate that our method reaches competitive results on both data sets, already with a single inference run with $K=1$. Additionally, for both data sets using the \gls{mmse} by averaging over multiple $K$ sampling runs gives a significant boost in quantitative performance. Moreover, using $T=100$ noise injection steps as described in Algorithm~\ref{alg:Inference} further improves the quantitative metrics, yielding state-of-the-art results.\\\\ 
Note that for \gls{ode}-based sampling with FlowSDF (i.e. $K=1$), the predictions for a single run seem to be closer to the \gls{mmse} than with \gls{sde}-based sampling. Hence, diffusion models experience a greater boost by increasing $K$ as observed in~\citep{amit2021segdiff}, whereas FlowSDF already exhibits high-quality results with $K=1$. 

\begin{figure}[h]
    \centering
    \includegraphics[width=\textwidth]{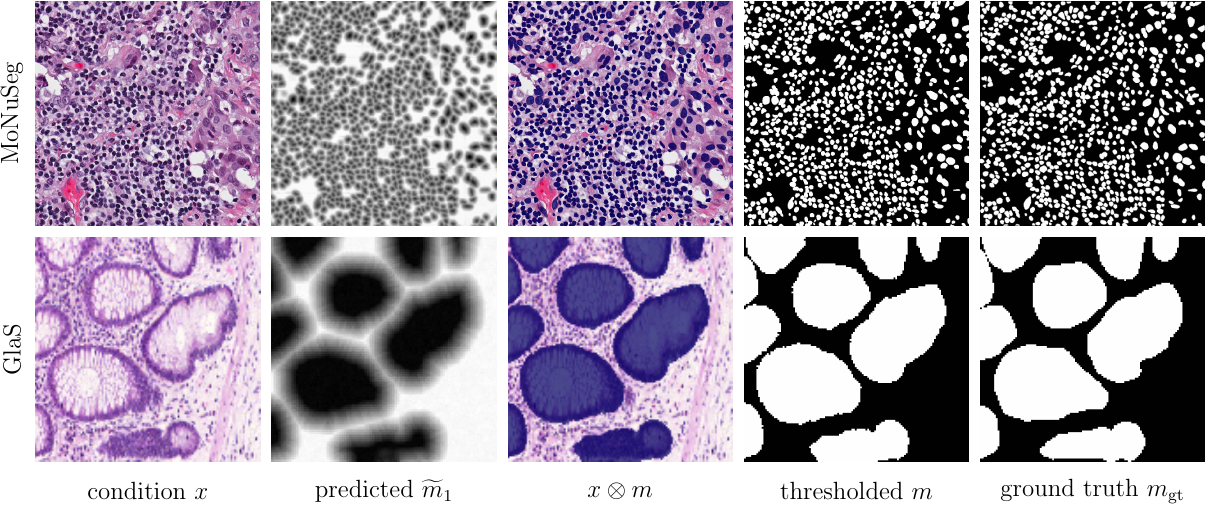}
    \caption{Exemplary sampled segmentation masks for both data sets. The predicted \gls{sdf} masks $\widetilde{m}_1$ are directly obtained from the \gls{ode} sampling, whereas the thresholded masks $m$ are shown to additionally enable a visual comparison with the depicted ground truth $m_{\mathrm{gt}}$. Furthermore, we provide the condition image with the overlaid thresholded mask $x\otimes m$.}
    \label{fig:results}
\end{figure}

\begin{figure}[h]
    \centering
    \includegraphics[width=\textwidth]{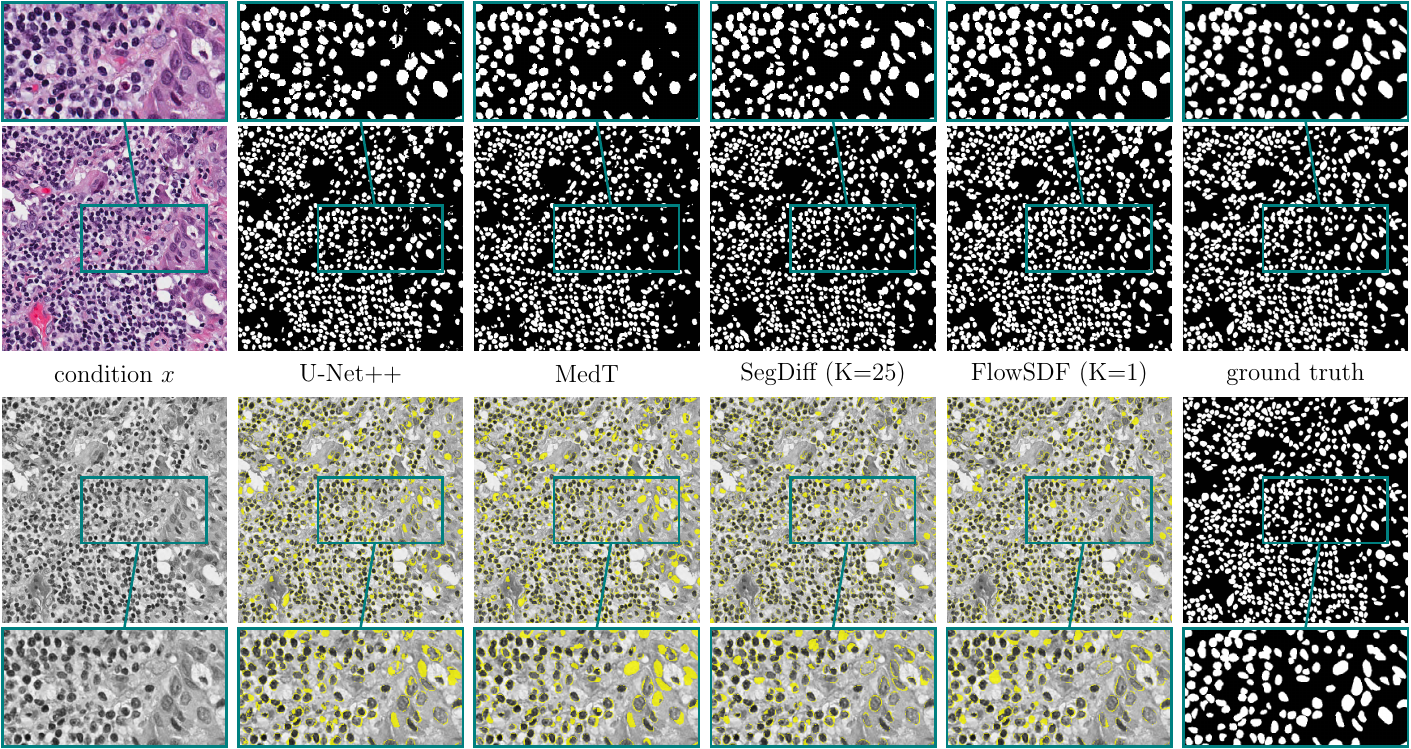}
    \caption{Qualitative results on a MoNuSeg test image $x$ for different selected discriminative and generative approaches. Top row: we compare FlowSDF to the generative \gls{ddpm} SegDiff~\citep{amit2021segdiff} and two discriminative models, U-Net++~\citep{zhou2018unet++} and MedT~\citep{valanarasu2021medical}. Bottom row: we present the respective difference images to the ground truth blended into the gray-scaled condition image for the approaches reported above. As depicted in the provided zoom, the generative methods yield smoother objects, which can be attributed to averaging over $K=25$ runs for Segdiff and to the \gls{sdf} with FlowSDF, where $K=1$ suffices.}
    \label{fig:results_visual_comp}
\end{figure}

\begin{table}[htb]
\caption{Quantitative segmentation results on the MoNuSeg and \gls{glas} data set. Best score for each metric in \textbf{bold}, second-best \underline{underlined}.}\label{tab:results}
\centering
\begin{tabular}{l|cc|cc} \hline
& \multicolumn{2}{c|}{MoNuSeg} & \multicolumn{2}{c}{\gls{glas}} \\ 
Method &  F1~$\mathrm{\uparrow}$ & m\gls{iou}~$\mathrm{\uparrow}$ &F1~$\mathrm{\uparrow}$ & m\gls{iou}~$\mathrm{\uparrow}$ \\ \Xhline{1.3pt}  
FCN~\citep{badrinarayanan2017segnet} & 28.84 &28.71 & 66.61 &50.84 \\
U-Net~\citep{ronneberger2015u} & 79.43 &65.99 & 77.78 & 65.34 \\
U-Net++~\citep{zhou2018unet++} & 79.49& 66.04&78.03 &65.55 \\
Res-UNet~\citep{xiao2018weighted} & 79.49 & 66.07 &78.83 &65.95 \\
Axial Attention U-Net~\citep{wang2020axial} & 76.83 & 62.49 &76.26& 63.03 \\
MedT~\citep{valanarasu2021medical} & 79.55 & 66.17 & 81.02& 69.61\\
SegDiff~\citep{amit2021segdiff} (K=1) &76.88 &62.59 & 86.40&77.36 \\
SegDiff~\citep{amit2021segdiff} -- \gls{mmse}~ (K=25) & \underline{81.19}& \underline{68.44} & 87.27 & 78.65 \\
\gls{ve} \gls{sde} with \gls{sdf}~\citep{bogensperger2023score} & 78.13 & 64.19& 82.03 & 71.36  \\ \hline
FlowSDF (K=1) & 80.67 & 67.70 & 88.64 & 80.63\\
FlowSDF -- \gls{mmse} (K=25) & 81.02 & 68.19 & \underline{88.94} &\underline{81.05} \\
FlowSDF -- \gls{mmse} (K=25,T=100)& \textbf{81.22}& \textbf{68.46}& \textbf{89.31} & \textbf{81.71} \\
\hline
\end{tabular} 
\end{table}
Figure~\ref{fig:results} shows exemplary qualitative results for both data sets to further highlight the potential applicability of FlowSDF. 
The segmented objects $m$ derived from thresholding the \gls{sdf} predictions $\widetilde{m}_1$ are distinctly cohesive and closely match the ground truth segmentation masks $m_{\mathrm{gt}}$ for both data set samples.
A visual comparison is additionally depicted in Figure~\ref{fig:results_visual_comp}, where we compare the thresholded segmentation prediction of FlowSDF for a MoNuSeg test image with the \gls{ddpm}-based (generative) model~\citep{amit2021segdiff} and two discriminative models, namely U-Net++~\citep{zhou2018unet++} and MedT~\citep{valanarasu2021medical}. As can be seen in the provided zoom, the \gls{sdf} representation of segmentation objects indeed seems to act as a shape prior and thus yields smoother segmentation objects while avoiding artefacts such as single pixels/small structures that are mistakenly classified as foreground objects. 
Note that SegDiff requires more runs ($K=25$ as reported in~\citep{amit2021segdiff}) in order to increase performance, while FlowSDF already yields comparatively good results
with one evaluation cycle ($K=1$), likely due to a combination of the \gls{sdf} representation and the \gls{ode}-based sampling.\\\\

\begin{figure}[h]
    \centering
    \includegraphics[width=\textwidth]{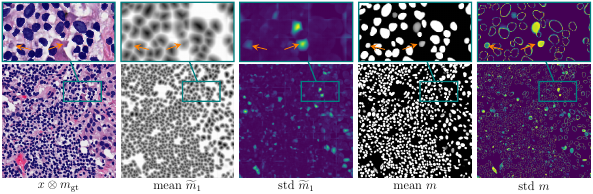}
    \caption{Segmentation example with according statistical values for the \gls{sdf} predictions $\widetilde{m}_1$ respectively thresholded masks $m$ (K=128).
    Notably, the image includes a region where erroneously two nuclei were segmented, as indicated by the orange arrows. This region is highlighted in the standard deviation maps, which represent the associated uncertainty in the segmentation.
    }
    \label{fig:uncertainty}
\end{figure}

\begin{figure}[h]
    \centering
    \includegraphics[width=\textwidth]{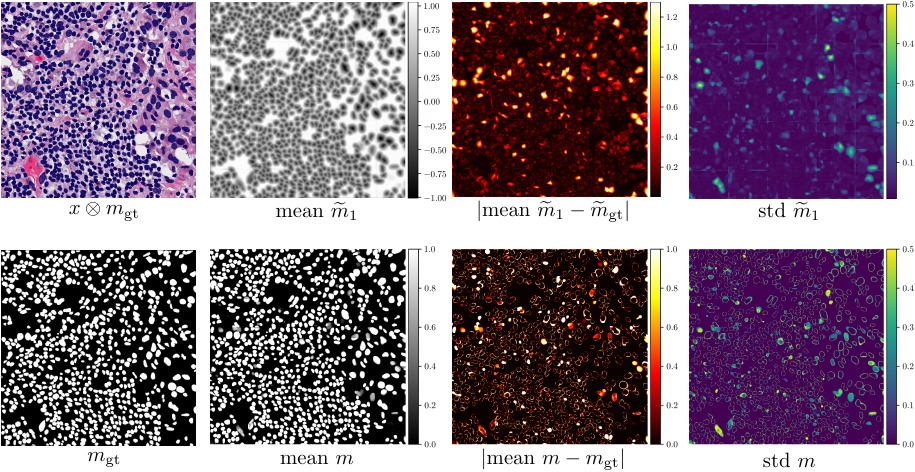}
    \caption{Segmentation example with statistical values for the predicted $\widetilde{m}_1$ and thresholded masks $m$, along with the corresponding ground truth images and the absolute error between the predictions and the ground truth (K=128).
    The visual comparison of error and standard deviation maps (uncertainty) suggests that the latter has predictive capability for the error.}
    \label{fig:uncertainty_error}
\end{figure}

Utilizing a generative model enables us to sample from the conditional distribution of the \gls{sdf} given the conditioning image $x$.
An advantage of this approach is given by the fact that the resulting statistical values such as the pixel-wise variance $\mathbb{V}[\widetilde{m}| x]$ allow for the quantification of segmentation uncertainties in the \gls{sdf} predictions as well as the thresholded masks, which provide additional insights into the segmentation process that are not available with traditional discriminative point estimators.
An illustration of the aforementioned property on a MoNuSeg test image $x$ can be seen in Figure~\ref{fig:uncertainty} for $K=128$, where the standard deviation maps associated with the \gls{sdf} predictions and thresholded masks highlight the regions of uncertainty. 
In general, we observe that the standard deviation appears high on transitions from nuclei to background as well as in wrongly detected nuclei or over-segmented parts of nuclei. 
This encouraging observation leads us to the hypothesis that the uncertainty may be associated directly with segmentation errors similar to what has been shown in \citep{narnhofer2022posterrquant}, see also Figure~\ref{fig:uncertainty_error}. Here, a visual comparison of the error and standard deviation (uncertainty) maps for the same test sample as in Figure~\ref{fig:uncertainty} indicates that the latter very likely has high predictive capability.  A detailed analysis including the mutual information of the two variables, however, is out of scope for this work. 

\subsection{Ablation}\label{sec:exp:ablation}
\begin{figure}[!htb]
    \centering
    \includegraphics[width=\textwidth]{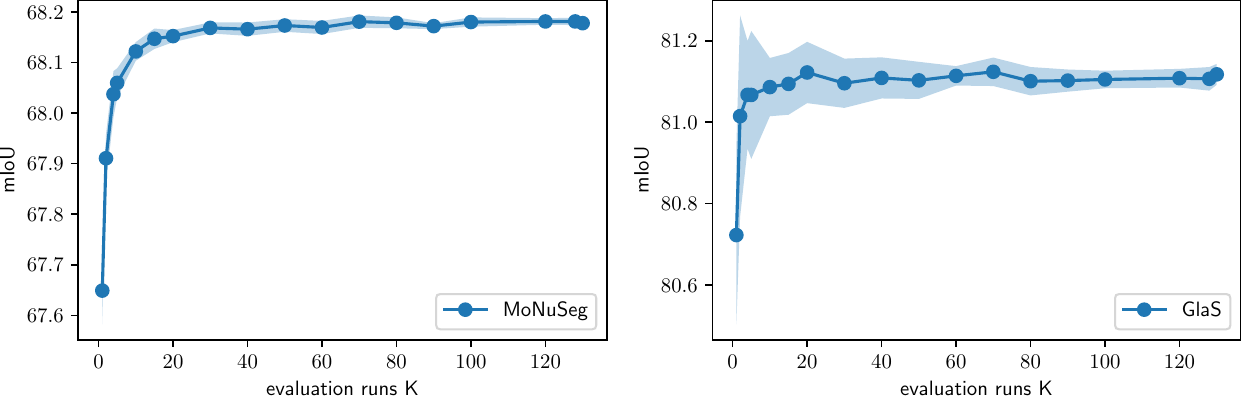}
    \caption{Influence of evaluation runs $K$ on mIoU averaged over 25 runs for both data sets. Note that \gls{ode}-based sampling initially benefits from increased runs $K$ but soon stabilizes and remains fairly constant.}
    \label{fig:ablation_K}
\end{figure}

We consider several ablation studies, namely
the effect of using \gls{sdf} representations for segmentation masks, the patch-averaging strategy to combine overlapping segments in the MoNuSeg test data and the number of ensemble runs $K$ when computing \gls{mmse} predictions. 
Table~\ref{tab:ablation_sdf} compares quantitative segmentation results for both data sets using binary masks (w/o \gls{sdf}) and FlowSDF (w \gls{sdf}), showing the performance increase due to using \glspl{sdf}. Note that the results are obtained using a single run with $K=1$, but the same trend is also observed using ensemble predictions.
\begin{table}[h]
\caption{Ablation results over \gls{sdf} representation compared to binary mask.}\label{tab:ablation_sdf}
\centering
\begin{tabular}{p{4.cm}|p{1.4cm} p{1.5cm} | p{1.4cm} p{1.5cm}} \hline
& \multicolumn{2}{c|}{MoNuSeg} & \multicolumn{2}{c}{\gls{glas}} \\ 
Method &  F1~$\mathrm{\uparrow}$ & m\gls{iou}~$\mathrm{\uparrow}$ &F1~$\mathrm{\uparrow}$ & m\gls{iou}~$\mathrm{\uparrow}$ \\ \Xhline{1.3pt}
w/o \gls{sdf} & 79.78 & 66.45  & 88.20 & 79.91\\
w \gls{sdf} & 80.67 & 67.70 &88.64 & 80.63\\
\hline
\end{tabular} 
\end{table}\\\\
Moreover, for evaluating test images of the MoNuSeg data set, overlapping patches of $128 \times 128$ as described in Section~\ref{sec:exp:eval} were extracted. The effect of averaging over its merged regions similar to~\citep{ozdenizci2023restoring} is shown in Table~\ref{tab:ablation_patch_merge}. Note that this is shown only for the MoNuSeg data set where the available images are too large to directly resize the whole images, as opposed to the \gls{glas} data set which allowed for evaluations on the entire images. 
\begin{table}[h]
\caption{Influence of patch-averaging strategy for overlapping segments in MoNuSeg.}\label{tab:ablation_patch_merge}
\centering
\begin{tabular}{p{4.cm}|p{1.4cm} p{1.5cm}} \hline
& \multicolumn{2}{c}{MoNuSeg} \\ 
Method &  F1~$\mathrm{\uparrow}$ & m\gls{iou}~$\mathrm{\uparrow}$ \\ \Xhline{1.3pt}
w/o patch-averaging & 80.45 & 67.38 \\
w patch-averaging & 80.67 & 67.70 \\
\hline
\end{tabular} 
\end{table}\\\\
Finally, the effect of $K$ independent sampling runs as described in Algorithm~\ref{alg:Inference} to approximate the \gls{mmse} is 
visualized in Figure~\ref{fig:ablation_K} for both data sets. We use only one noise injection step $T$ here, to recover the original \gls{ode}-based sampler as described in~\citep{lipman2022flow}. It is noteworthy that unlike \gls{sde}-based samplers that are used in diffusion models, \gls{ode} samplers seem to better approximate the \gls{mmse} already for smaller values of $K$. Thus, the performance gain is not as large although increasing $K$ moderately definitely helps in boosting quantitative metrics.

\section{Conclusion}
In this work we proposed FlowSDF, a generative approach for medical image segmentation by fusing image-guided flow matching with the concept of representing segmentation maps with the \gls{sdf}. 
The applicability of FlowSDF was demonstrated qualitatively and quantitatively on two public medical data sets. The \gls{sdf} provides a smoother and more continuous representation of object boundaries compared to binary masks, which makes it a more canonical choice for accurate and robust segmentation.
Moreover, by leveraging the generative approach, statistical measures such as mean and standard deviation can be computed to quantify the uncertainty in the segmentation results. 
This information is especially useful for medical diagnosis and treatment planning, where it is important to know the level of confidence in the segmentation results.\\
In the future, it is of interest to investigate the benefit of the \gls{sdf} in object detection and segmentation beyond the medical domain where data sets contain more complex shapes that are in general non-circular. The extension to multi-class semantic segmentation could also be studied. 
Finally, with the success of \glspl{ldm}~\citep{rombach2022high}, the usage of foundation models should be investigated. As \glspl{ldm} have also been developed in the medical domain~\citep{yellapragada2024pathldm}, it is of high interest to study the extent to which they can capture semantic concepts and potentially improve upon existing methods. 
With the outlook of a recent, powerful foundational model that is based on flow matching~\citep{esser2024scaling}, their usage in a conditional setting for image segmentation can be explored. Such approaches have already been successfully applied for learning conditional distributions in depth estimation~\citep{gui2024depthfm,ke2023repurposing}.

\newpage
\bibliographystyle{unsrtnat}
\bibliography{main}

\end{document}